\documentclass[10pt,journal,compsoc]{IEEEtran}

\usepackage{amsthm}
\usepackage{booktabs}
\usepackage{array}
\usepackage{hyperref}
\usepackage{overpic}
\usepackage{amssymb}
\usepackage{graphicx}
\usepackage{epstopdf}
\usepackage[english]{babel}
\usepackage[numbers]{natbib}
\usepackage{bm}
\usepackage{float}
\usepackage{algorithm}
\usepackage{multirow}
\usepackage{longtable}
\usepackage{subfigure}
\usepackage{subcaption}
\usepackage{supertabular}
\usepackage{algpseudocode}
\usepackage{amsmath}
\usepackage[justification=centering]{caption}
\DeclareMathOperator*{\argmax}{argmax}
\usepackage{algorithm,float}
\makeatletter

\ifCLASSINFOpdf

\fi

\iffalse
\else

\begin{document}

\title{Interpretable Multi-View Clustering}

\author{Mudi Jiang, Lianyu Hu, Zengyou He,  Zhikui Chen  
\IEEEcompsocitemizethanks{
\IEEEcompsocthanksitem M. Jiang, L. Hu and Z. He are with School of Software, Dalian
University of Technology, Dalian, China.\protect\\

\IEEEcompsocthanksitem   Z. Chen (corresponding author) is with School of Software, Dalian University of Technology, Dalian,
China, and Key Laboratory for Ubiquitous Network and Service Software
of Liaoning Province, Dalian, China.\protect\\ Email: zkchen@dlut.edu.cn
}
\thanks{Manuscript received XXXX 2024; revised XXXX, 2024.}}

%
%

\markboth{IEEE Transactions on Knowledge and Data Engineering,~Vol.~XX, No.~XX, May~2024}%
{Shell \MakeLowercase{\textit{et al.}}: Bare Advanced Demo of IEEEtran.cls for IEEE Computer Society Journals}

\IEEEtitleabstractindextext{%
\begin{abstract}
Multi-view clustering has become a significant area of research, with numerous methods proposed over the past decades to enhance clustering accuracy. However, in many real-world applications, it is crucial to demonstrate a clear decision-making process—specifically, explaining why samples are assigned to particular clusters. Consequently, there remains a notable gap in developing interpretable methods for clustering multi-view data. To fill this crucial gap, we make the first attempt towards this direction by introducing an interpretable multi-view clustering framework. Our method begins by extracting embedded features from each view and generates pseudo-labels to guide the initial construction of the decision tree. Subsequently, it iteratively optimizes the feature representation for each view along with refining the interpretable decision tree. Experimental results on real datasets demonstrate that our method not only provides a transparent clustering process for multi-view data but also delivers performance comparable to state-of-the-art multi-view clustering methods. To the best of our knowledge, this is the first effort to design an interpretable clustering framework specifically for multi-view data, opening a new avenue in this field.
\end{abstract}

}

\maketitle

\IEEEdisplaynontitleabstractindextext

\IEEEpeerreviewmaketitle

\ifCLASSOPTIONcompsoc
\IEEEraisesectionheading{\section{Introduction}\label{1}}

Cluster analysis \cite{romesburg2004cluster,saxena2017review} stands as a pivotal task in the realm of data mining. Conventional clustering methods (e.g. $k$-means \cite{macqueen1967some}) are primarily designed for single-view data. As the data landscape evolves to feature information from multiple perspectives and sources, the data sets that need to be analysed become increasingly complex. This complexity has brought multi-view clustering (MVC) \cite{chao2021survey,fu2020overview,wang2019multi} to the forefront in recent years, as it effectively integrates diverse information and provides a deeper understanding of complex data.

Existing MVC methods can be roughly categorized into four subgroups: subspace methods, graph-based methods, matrix factorization methods and deep learning methods. While the models proposed for MVC have already achieved high clustering accuracy, how to explain the reported clusters is still an underlooked  issue so far. In general, the interpretability refers to the capability of  enabling people to understand  how the clustering results are derived.  A common practice is to characterizing the clustering result in terms of a decision tree or a set of rules.  However, existing MVC methods generally exhibit a significant shortfall in terms of interpretability. This deficiency hinders their practical usage, as users often struggle to understand the logic and reasoning behind the clustering outcomes, making it challenging to fully trust and utilize these methods in real-world scenarios.

Interpretable clustering \cite{bandyapadhyay2023find,bertsimas2021interpretable} has garnered significant attention in recent years. The most widely acknowledged  interpretable algorithms employ decision trees or sets of rules,  clearly illustrating why an instance is assigned to a cluster.  However, to  our best knowledge, existing interpretable clustering methods predominantly focus on single-view data. Consequently, there remains a notable gap in research specifically addressing multi-view clustering, highlighting an area ripe for further exploration.

Motivated by the aforementioned observations,  we propose an interpretable multi-view clustering framework,  which aims to concurrently refine multi-view feature representations and an interpretable decision tree. Initially, pseudo-labels derived from embedded features  guide the construction of the decision tree.  The model's interpretability and accuracy are subsequently enhanced through a joint optimization framework that proceeds as follows: (1) the decision tree structure is fixed while the embedded features are optimized, improving the quality of the pseudo-labels. (2) fixing the embedded features to fine-tune the decision tree, thereby boosting its interpretability. This bi-phasic   optimization  fosters a potent  synergy between feature representation and decision-making clarity, establishing the cornerstone of our interpretable multi-view clustering framework.


To validate the efficacy of our proposed framework, we conduct a series of experiments on various benchmark data sets. The experimental results indicate that the proposed method  is not only comparable to the current state-of-the-art (SOTA) MVC methods with respect to the clustering quality but also retains a high degree of interpretability. Furthermore, our method significantly outperforms the existing interpretable clustering methods designed for single-view data.

In summary, the main contributions of this paper are outlined as follows: 
\begin{itemize}
    \item We present a novel MVC algorithm, pioneering the integration of interpretability into the MVC field. This unique approach opens a new avenue in MVC research, particularly in enhancing the interpretability of MVC algorithms.
    \item A joint optimization clustering framework has been devised, which iteratively refines the embedded feature representations and the decision tree. Such a framework  not only improves the clustering accuracy but also enhances the model's interpretability, making the clustering outcomes more transparent and trustworthy.
    \item Experimental results on benchmark data sets demonstrate that the proposed method  not only maintains competitive performance compared  with SOTA  MVC methods but also surpasses other existing interpretable single-view clustering methods.
\end{itemize}

The rest of this paper is organized as follows.
Section \ref{2} gives a discussion on the related work. Section \ref{3} provides a detailed description about the proposed framework. Section \ref{4} presents the experimental results. Section \ref{5} concludes the paper and discusses  future work.

\section{Related work}
\label{2}
\subsection{Multi-view Clustering}
Each category of MVC methods, as outlined in Section \ref{1}, employs its own set of strategies to navigate the complexities inherent in multi-view data. Subspace methods primarily focus on discovering  latent shared subspaces within data from multiple views, followed by conducting cluster analysis within these identified subspaces \cite{li2019flexible,li2019reciprocal,yang2019split}. Graph-based  methods typically involve constructing a graph for each view where nodes represent data points and edges reflect similarities or relationships \cite{liang2019consistency,li2021consensus}. Matrix factorization methods decompose data from each view into lower-dimensional matrices, uncovering the latent structures that characterize the inherent relationships and patterns within the multi-view data \cite{wang2018multiview,yang2020uniform}.  Given the  relevance of deep learning methods \cite{li2021adaptive,xu2022multi} to the framework proposed in this paper, a concise overview of these methods is provided in the following subsection.
\subsubsection{Two-stage deep learning methods}
Two-stage  deep learning methods for MVC employ a sequential process \cite{wang2015deep}. Initially,  deep neural networks  are deployed in the first stage for feature extraction, effectively learning complex representations from the data. The subsequent stage capitalizes on these learned features, employing traditional clustering algorithms like $k$-means or spectral clustering \cite{li2019deep,gao2020cross} to perform the partitioning process. This approach distinctly delineates the feature learning phase from the clustering phase, permitting each stage to be finely tuned and executed independently.
\subsubsection{One-stage deep learning methods}
In one-stage methods,  feature learning and clustering tasks are simultaneously optimized within a unified framework. The most widely employed framework is derived from deep embedded clustering (DEC) \cite{xie2016unsupervised}, which utilizes a deep stacked autoencoder. Following this, the model is iteratively optimized, focusing on a clustering objective based on Kullback-Leibler (KL) divergence, in conjunction with a self-training target distribution.   To tackle multi-view clustering tasks, the aforementioned framework is expanded in \cite{xu2022self,xie2020joint}, employing distinct autoencoders for each view to generate view-specific cluster assignments and a unified distribution for all views. These methods iteratively mine latent features and refine clustering structure by optimizing a combination of reconstruction and KL divergence losses. 
\subsection{Interpretable Clustering}
Interpretable clustering methods can be categorized into various groups based on the models utilized to elucidate the assignment of instances to different clusters.  These methods encompass approaches such as \textit{if-then} rules \cite{balachandran2009interpretable}, polytopes \cite{lawless2022interpretable} and hyperrectangles \cite{chen2016interpretable}. Given that this paper employs a binary decision tree as the interpretable model, relevant literature and methodologies are delineated in the subsequent subsections.
\label{2.2}
\subsubsection{Two-stage tree construction}
Two-stage methods for developing interpretable clustering trees typically begin by utilizing conventional clustering techniques to generate pseudo-labels, followed by the construction of a supervised decision tree \cite{bandyapadhyay2023find,gabidolla2022optimal}. The process of selecting the split feature value at each internal node can be approached in various ways. For instance, the method outlined in  \cite{dasgupta2020explainable} aims to minimize the number of misclassified nodes at each split. Alternatively, the approach described in \cite{makarychev2022explainable}  involves identifying the median of all cluster centers associated with a node and subsequently calculating the maximum distance from these centers to the median.  Furthermore, a joint optimization framework, presented in \cite{gabidolla2022optimal}, alternately optimizes variables learned by clustering algorithms (such as the cluster centroids in $k$-means) and the parameters of the tree.
\subsubsection{One-stage tree construction}
 In contrast to the two-stage tree construction approaches, an interpretable clustering tree can be directly constructed by leveraging the inherent information within the data. In \cite{basak2005interpretable}, four different measures are presented to select the most appropriate attribute, which are used to split the data at every internal node. The algorithm proposed in \cite{fraiman2013interpretable,ghattas2017clustering}  focuses on minimizing heterogeneity within each node, thereby enhancing uniformity. Empirical assessments of probabilities and deviations guide the selection of variables and split points, aiming at a significant reduction of variance within each node for more accurate data clustering.
\section{Method}
\label{3}
\begin{figure*}[htbp]
	\includegraphics[scale=0.6]{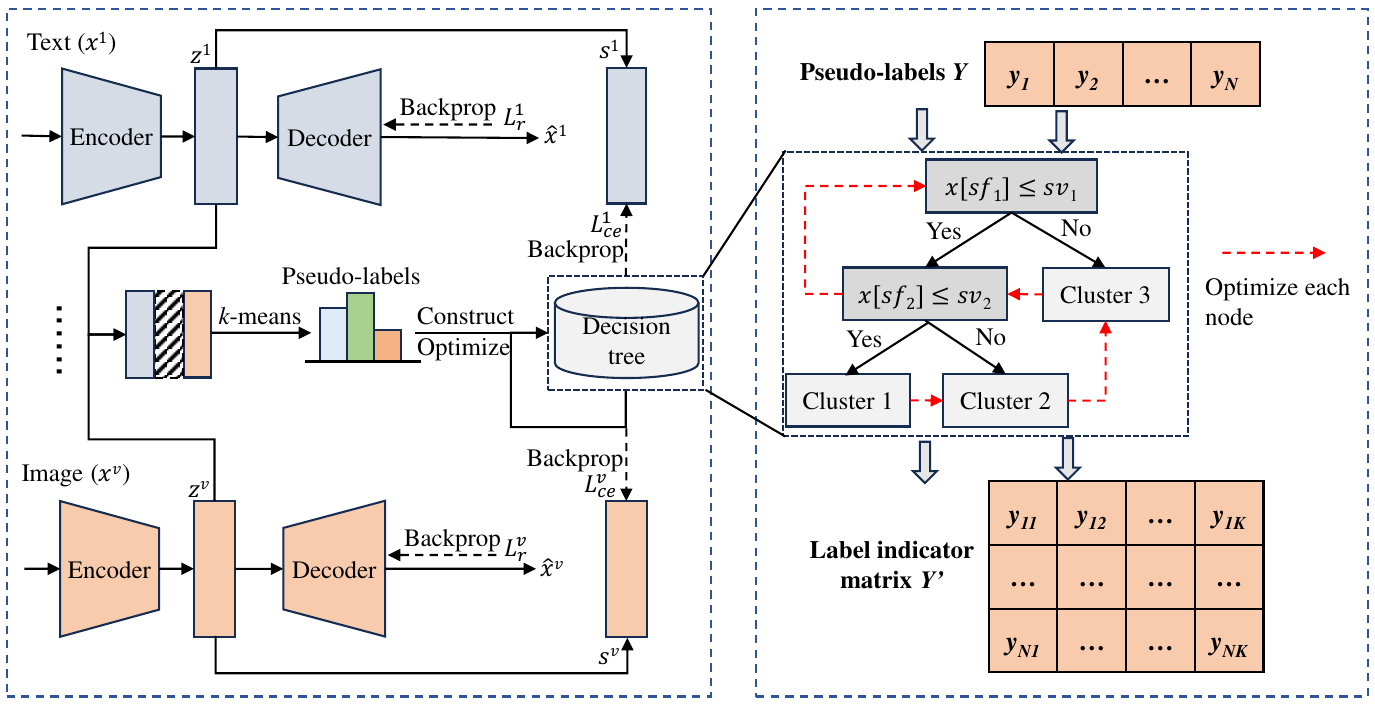}  
	\centering
	\caption{The joint optimization framework for interpretable MVC.}  
	\label{fig1}   
\end{figure*}
\begin{table*}[htbp]
	\caption{Notations.}
	\label{table1}
	\centering
	\begin{tabular}{cc}
		\toprule
		Notation &  Definition \\
		\midrule
            $N,V,K$ & Number of instances, views, clusters\\
            $R_v$ & Dimensionality of the $v$th view \\
            $E=(E^1,...,E^V)$ &  $V$ encoders \\
            $D=(D^1,...,D^V)$ &  $V$ decoders \\
		$\theta=(\theta^1,...,\theta^V)$ &  Parameters of $V$ encoders \\
            $\phi=(\phi^1,...,\phi^V)$ & Parameters of $V$ decoders \\
            $e_1$,$e_2$ &  Number of training epochs    \\
            $Z = \{z_i^v \mid 1 \leq i \leq N; 1 \leq v \leq V\}$ & Concatenated embedded features \\
            $Y=\{y_i\mid 1 \leq i \leq N\}$ & Pseudo-labels\\
            $Y'= \{y_{ij} \mid 1 \leq i \leq N; 1 \leq j \leq K\}$ & Label indicator matrix\\
            $S = \{s_{ij}^v \mid 1 \leq i \leq N; 1\leq j \leq K; 1 \leq v \leq V\}$ &  Soft assignment \\
            $\mathbb{I}(,)$ & Indicator function, 1 if arguments are equal, 0 otherwise\\
            $T$ & Decision tree\\
            $b_i$ & A node\\
            $sf_i, sv_i$ & Split feature and value of $b_i$\\
            $T(x_n; SF,SV)$ & Output label of $x_n$ via the decision
            tree\\
            $t(x_n; sf_i,sv_i)$ & Output label of $x_n$ of the
subtree with root at node $b_i$\\
            $l_i$ & Label of  $b_i$\\
            \bottomrule
	\end{tabular}
\end{table*}

The multi-view clustering task is to partition a collection of $N$ instances, denoted as $\{x_{i}^{v} \in \mathbb{R}^{R_v}\}_{i=1}^{N}$, into $K$ distinct clusters. Here, $v$ represents the $v$th view and $R_v$ signifies the dimensionality of that view.  Other symbols employed throughout this paper and their corresponding definitions are presented in Table \ref{table1}. The proposed framework, composed of two key steps: model initialization and model optimization, is shown in Fig. \ref{fig1}.
\begin{itemize}
    \item Initially, distinct autoencoders are pre-trained for each view to mine the embedded features. Subsequently, features from all views are  concatenated to generate pseudo-labels by employing $k$-means. A standard decision tree is then constructed in the original feature space, guided by these pseudo-labels.
    \item The model optimization process is executed iteratively. In the first phase,  outputs from the decision tree's leaf nodes are utilized to compare  with the view-specific cluster assignments, facilitating the refinement of the autoencoders. In the subsequent phase, the re-concatenated features are employed to generate the initial set of refined pseudo-labels. These initial labels facilitate the commencement of the decision tree's self-iterative optimization process, in which the decision tree leverages its own outputs, from the preceding iteration, as the new set of pseudo-labels for continuous refinement.  The process of self-generated label optimization persists until the decision tree's structure achieves a state of stability, with no further modifications observed. This constitutes a complete cycle of joint optimization iteration.
\end{itemize}

\subsection{Model Initialization}
The process for initializing the model, involving the pre-training of autoencoders and the construction of a decision tree, is summarized in Algorithm 1.
\begin{algorithm}[htbp]
	\small
	\caption{Model initialization}
	\begin{algorithmic}[1] 
		\Require Multi-view data set $\mathcal{X}$, number of clusters $K$, number of epochs $e_1$, max depth of the decision tree $maxDep$, minimum number of instances in a node $minNum$.
		\Ensure A set of pre-trained autoencoders, a decision tree.
            \Function {Pre-train autoencoders}{$e_1$}
            \State Let $ite=1$
            \While{$ite \leq e_1$}
            \State Compute reconstruction loss $L^v_r$ for all views using  Eq. (3) 
            \State Update network parameters $\theta^v$, $\phi^v$ using adaptive  moment estimation
            \State $ite++$
            \EndWhile
            \State\Return $E^v,D^v$
            \EndFunction
            \State Obtain embedded features $z_i^v$ using Eq. (1)
            \State Apply $k$-means on concatenated features $Z$ to obtain pseudo-labels $\{Y=y_i\}_{i=1}^{N}$
            \Function {Build tree}{$\mathcal{X},minNum,maxDep,Y$}
		\If{$Terminal(\mathcal{X},minNum,maxDep,Y)$}
		\State  $T=leaf(\mathcal{X})$
	    \State\Return $T$
		\EndIf
            \State $\mathcal{X}_L,\mathcal{X}_R=best\_spilt(\mathcal{X},sf,sv)$
		\State $T.left=$ \Call{Build tree}{$\mathcal{X}_L,minNum,maxDep,Y$}
		\State $T.right=$\Call{Build tree}{$\mathcal{X}_R,minNum,maxDep,Y$}
            \State\Return $T$
		\EndFunction
	\end{algorithmic}
\end{algorithm}

\subsubsection{Pre-train AE}
To efficiently extract the intrinsic information contained within the data, a set of deep autoencoders are employed to ascertain  latent representations for each distinct view. More precisely, for the $v$th encoder $E^v$ characterized by the parameters $\theta^v$, the embedded features of the corresponding view are derived as follows:
\begin{equation}
    z_i^v=E^v(x_i^v;\theta^v).
\end{equation}
Similarly, the output of $v$th decoder $D^v$ is expressed as 
\begin{equation}
    \hat{x}_i^v=D^v(z_i^v;\phi^v)=D^v(E^v(x_i^v;\theta^v);\phi^v),
\end{equation}
where $\phi^v$ denotes the parameters of $D^v$. The reconstruction loss, quantifying the discrepancy between the output of the autoencoders $\hat{x}_i^v$ and the original input data $x_i^v$, is defined as
\begin{equation}
    L^v_r=\sum_{i=1}^{N} {\parallel \hat{x}_i^v-x_i^v \parallel}^2,
\end{equation}
and is minimized through backpropagation to refine the feature representations of each view for subsequent processing stages (lines 3$\sim$7).
\subsubsection{Construct decision tree}
In this phase, we initially produce a set of pseudo-labels by implementing 
$k$-means clustering on the global features, which are concatenated from the feature representations across all views (lines 10$\sim$11):
\begin{equation}
    Z_i=[z_i^1;z_i^2,,,;z_i^V].
\end{equation}
Utilizing these pseudo-labels, a standard decision tree is constructed in a supervised manner (line 12). 
All samples originate at the root node, which is iteratively partitioned into two child nodes through the following process: (1) identifying the optimal split feature $sf \in \mathbb{R}^{\sum_{v=1}^{V}R_v}$ and spilt value $sv$. (2) allocating instances to the left or right child node contingent upon whether $x_i[sf] \leq sv$ holds. Within the $best\_split$ function (line 17), the selection of the optimal feature and its split value is accomplished by exhaustively evaluating all features and corresponding values of the instances present in the node.  The splitting criterion's efficacy is measured by the Gini index, which for a set $\mathcal{X}$ is articulated as:
\begin{equation}
    Gini(\mathcal{X})=1-\sum_{i=1}^{K}p_i(\mathcal{X})^2,
\end{equation}
where $K$ represents the number of clusters, and $p_i(\mathcal{X})$ is the portion of instances in the $i$th cluster within $\mathcal{X}$.  The discriminative capacity of a chosen split point for node $\mathcal{X}$  is determined by:
\begin{equation}
	Gini(\mathcal{X},sf,sv)=\frac{\left|\mathcal{X}_L \right| }{\left|\mathcal{X} \right|}Gini(\mathcal{X}_L)+\frac{\left|\mathcal{X}_R \right| }{\left|\mathcal{X} \right|}Gini(\mathcal{X}_R),
\end{equation}
where $\left|\mathcal{X}_L \right|$ and $\left|\mathcal{X}_R \right|$ indicate the counts of instances in the left and right child nodes, respectively.  
The recursive splitting is terminated when any of the following criteria are met: (1) the number of instances at the current node falls below a minimum threshold $minNum$, (2) the depth of the tree reaches $maxDep$, (3) all instances within the node have the same label (lines 13$\sim$15).

\textbf{Time complexity analysis.}
Let $N$, $V$, $L$ and $M$ represent the number of instances, views, layers in the autoencoders, and the maximum number of neurons in any layer, respectively. The computational complexity of processing a single sample through $V$ autoencoders is $\mathcal{O}(V \cdot L \cdot M^2)$. Assuming the number of training epochs in this phase is $e_1$, the time complexity for pre-training the autoencoders is $\mathcal{O}(e_1 \cdot N \cdot V \cdot L \cdot M^2)$. The construction of a standard decision tree incurs a time complexity of $O({\sum_{v=1}^{V}R_v} \cdot{N} \cdot \log_2{N})$ \cite{sani2018computational}.
\subsection{Model Optimization}
The model optimization process is implemented through an alternating optimization approach. Specifically, in each iteration, we  fix the decision tree $T$ to update the parameters $\theta^v$ and $\phi^v$. Subsequently, with $\theta^v$ and $\phi^v$ fixed, we proceed to optimize the decision tree $T$.
\begin{algorithm}[htb]
	\small
	\caption{Model optimization}
	\begin{algorithmic}[1] 
		\Require A set of pre-trained autoencoders $E^v$ and $D^v$, a decision tree $T$, number of epochs $e_2$, trade-off coefficient $\lambda$.
		\Ensure Autoencoders $E^v$ and $D^v$ with updated parameters, an optimized decision tree.
            \Function {Optimize feature representation} {}
            \State Let $ite=1$
            \While{$ite \leq e_2$}
            \State Compute  view-specified soft assignment $S^v$ using Eq. (7)
            \State Compute  $L^v$ for all views using  Eq. (8) 
            \State Update network parameters $\theta^v$, $\phi^v$ using adaptive  moment estimation
            \State $ite++$
            \EndWhile
            \State\Return $E^v,D^v$
            \EndFunction
            \State Obtain embedded features $z_i^v$ using Eq. (1)
            \State Apply $k$-means on concatenated features $Z$ to obtain pseudo-labels $\{Y=y_i\}_{i=1}^{N}$
            \Function {Tree optimization}{$T,Y$}
            \Repeat 
            \For{node $b_i \in  T$ visited in reverse breadth-first traversal} 
            \If{$b_i$ is leaf node}
            \State $l_i = \argmax_y \sum_{x_n \in b_i} \mathbb{I}(y_n, y)$
            \Else
            \State Optimize parameters of $b_i$ using Eq. (12)
            \EndIf
            \EndFor
            \State Prune empty nodes
            \State Reallocate instances to leaf nodes
            \Until{The structure of $T$ is no longer changed}
            \State\Return Optimized decision tree $T$
		\EndFunction
	\end{algorithmic}
\end{algorithm}

\subsubsection{Optimize feature representation}
The efficacy of the constructed decision tree is intricately linked to the quality of pseudo-labels. Therefore, enhancing the quality of these labels within each iteration becomes critical. We achieve this objective by integrating the common outputs across views (decision tree) with the distinctive information inherent to each view (lines 3 $\sim$ 8).

The decision tree $T$ facilitates the generation of a consistent label distribution ${Y' = y_{ij}}$ through its leaf nodes, signifying that instance $x_i$ is assigned to the $j$-th cluster.  This set of labels is utilized to benchmark against the view-specific cluster assignments, which are derived based on Student's $t$-distribution \cite{van2008visualizing}.  For a given view $v$, the soft cluster assignment (probability) that instance $x_i$ belongs to the $j$-th cluster is determined by the equation:
\begin{equation}
    s_{ij}^v=\frac{(1 + \| z_i^v - c_j^v \|^2)^{-1}}{\sum_{j} (1 + \| z_i^v - c_{j}^v \|^2)^{-1}},
\end{equation}
where $c_{j}^v$ denotes the center of the $j$-th cluster for view $v$. To ensure comprehensive learning of the intrinsic information across multi-view data, each view's autoencoder parameters are independently optimized through a dual objective comprising reconstruction loss and cross-entropy loss. The latter measures the discrepancy between the view-specific cluster assignment $s_{ij}^v$ and the consistent label distribution $Y'$:
\begin{equation}
    L^v=L_{r}^{v}+ \lambda L_{ce}^{v},
\end{equation} 
where $\lambda$ is the trade-off coefficient. The cross-entropy loss is articulated as:
\begin{equation}
    L_{ce}^{v}= - \sum_{i} \sum_{j} y_{ij} \log(s_{ij}).
\end{equation}
By optimizing the combined loss function, we update the parameters $\theta^v$ and $\phi^v$ for each view, enabling the acquisition of refined pseudo-labels through cluster analysis on the re-embedded features $Z$. This enhances the foundation for the decision tree's subsequent optimization process with higher-quality labels.
\subsubsection{Optimize decision tree}
In this phase, we employ an iterative self-optimization process for the decision tree, aiming to improve its quality and reduce its size, thereby increasing interpretability.

With a decision tree of fixed structure, a clear optimization objective is to minimize the total misclassification cost across all leaf nodes, characterized by parameters $SF,SV=\{sf_i, sv_i\}$ for each node $b_i$:
\begin{equation}
    L_T=N-\sum_{n=1}^{N} \mathbb{I}(y_n, T(x_n; SF,SV)),
\end{equation}
where the indicator function  $\mathbb{I}(y_n, T(x_n; SF,SV))=1$ if the instance $x_n$ reaches a leaf node with a matching true label $l_i$ via the decision path from the root, and $\mathbb{I}(y_n, T(x_n; SF,SV))=0$ otherwise. Based on the separability condition theorem proven in \cite{carreira2018alternating}, the overall objective function can be decomposed into two parts: (1) instances traversing an internal node $b_i$ and (2) remaining instances passing through nodes that are independent of the former and mutually independent as well. Hence, Equation (10) is reformulated as follows:
\begin{align}
L_T = &N-\sum_{x_n \in b_i} \mathbb{I}(y_n, T(x_n; sf_i, sv_i)) \nonumber \\
& - \sum_{x_n \in \mathcal{X} \setminus b_i} \mathbb{I}(y_n, T(x_n; s f_i, s v_i)),
\end{align}
which indicates that the optimization problem of a decision tree can be formulated as a series of smaller, independent problems for individual nodes. In each depth, nodes are independent from each other, so we adopt a reverse breadth-first traversal strategy for optimizing each node within the decision tree.

The optimization of an internal node aims to minimize the misclassification of instances that reach it. This objective reduces to a simplified problem: minimizing the binary misclassification loss for a particular subset of instances that arrive at the node (lines 15 $\sim$ 23):
\begin{equation}
    \min_{sf_i,sv_i} {-\sum_{x_n \in b_i} \mathbb{I}(\hat{y}_n, t(\hat{x}_n; sf_i,sv_i))},
\end{equation}
where $t(\hat{x}_n; sf,sv)$ represents the predicted label of the subtree with root at node $b_i$. Here, $\hat{x}_n$ signifies the subset of instances that are channeled to the node, which will be further clarified subsequently.

With the parameters of other internal nodes held constant, the determination of the final leaf node an instance arrives at is solely based on the child node (left or right) it follows. Therefore, the node optimization issue can be solved by enumerating and evaluating split features and values to identify the optimal split that minimizes binary misclassification cost. It is important to note that when an instance is forwarded to the left child node based on a chosen split feature and value, we encounter four possible outcomes: (1) The instance is labeled correctly at the leaf node, regardless of being in the left or right child node. (2) The instance is labeled incorrectly at the leaf node, regardless of the node side. (3) The instance is only labeled correctly in the left child node, not the right. (4) The instance is incorrectly labeled in the left child node but would be correct in the right. Since altering the decision function in the first two cases does not affect the outcome for the instances, these instances are excluded from the computation of our objective function.

To illustrate the computation of the objective function for an internal node, we present a toy example in Fig. \ref{fige}. Consider a scenario where six instances, each with one of three pseudo-labels (represented by three colors in the table), need to be clustered. The current decision tree is depicted to the left of the dashed line. Assuming optimization is required for one of the internal nodes (indicated by the red border), we first compute the objective function based on the current splitting feature and value. Out of the five samples reaching this node, $x_2$ and $x_5$ are mistakenly assigned to the wrong leaf nodes. However, the assignment of $x_2$ cannot be corrected by any change in the splitting feature or value of this node implying that $x_2$ does not contribute to evaluating the discriminative power of the splitting feature and value at this node, and thus is not considered. Consequently, the objective function for this node is 1, as only $x_5$  is misassigned. When we experiment with different splitting conditions (as shown to the right of the dashed line), $x_5$  is accurately assigned to its corresponding leaf node, reducing the objective function to 0. This successful reduction prompts the replacement of the splitting condition.
\begin{figure}[htbp]
	\includegraphics[scale=0.52]{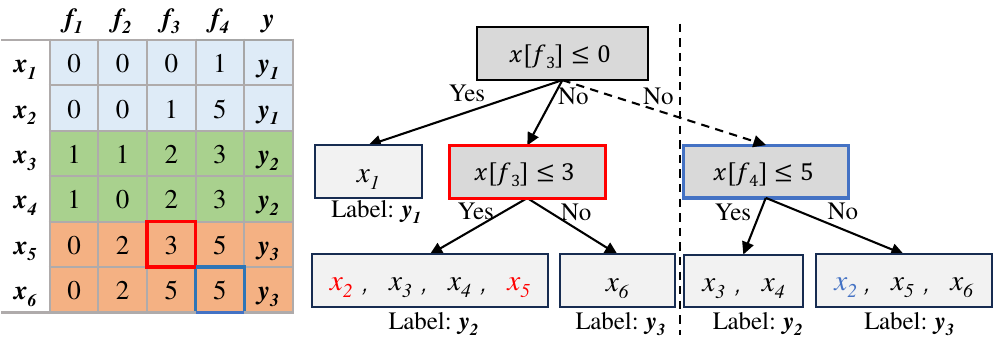}  
	\centering
	\caption{Illustration of objective function computation for an internal node in a decision tree. The figure shows the initial node configuration to the left of the dashed line. The right side of the dashed line demonstrates the effect of altering the splitting condition.}  
	\label{fige}   
\end{figure}

The optimization of the entire decision tree is carried out iteratively, using the output labels from the previous iteration as the input for the current iteration, and this process continues until there are no further changes in the tree's structure. Note that at the end of each iteration, we prune the empty nodes within the tree and reallocate each instance to its leaf node according to the updated nodes' parameters (line 22 $\sim $ 23). 

\textbf{Time complexity analysis.}
 The time complexity for computing the view-specific cluster assignment across all views is $O(N \cdot \sum_{v=1}^{V}R_v \cdot{K})$. Assuming the process undergoes $e_2$ epochs, the overall time complexity becomes $\mathcal{O}(e_2 \cdot N \cdot V \cdot L \cdot M^2)$.
Regarding the optimization of the decision tree, the complexity for a single iteration, which involves traversing all nodes, is comparable to the complexity of constructing a decision tree of equivalent size. Denoting $I$ as the average number of such iterations, the time complexity for this stage is quantified as $O({I \cdot \sum_{v=1}^{V}R_v} \cdot{N} \cdot \log_2{N})$.

\section{Experiments}
\label{4}
In this section, we conduct  a series  of experiments to evaluate the performance of the proposed method. These experiments were carried out on a PC with an Intel(R) Core(TM) i7-10700F CPU at 2.90 GHz, 16 GB RAM, and a GeForce RTX 1660 GPU with 6 GB of memory.
\subsection{Datasets}
We utilize the following five benchmark datasets in our experiments:
\begin{itemize}
    \item \textbf{Mfeat} \footnote{https://archive.ics.uci.edu/dataset/72/multiple+features}: This dataset comprises 2000 handwritten numerals ('0'-'9') sourced from Dutch utility maps. Each numeral is described by six feature sets: 76-dimensional FOU, 216-dimensional FAC, 64-dimensional KAR, 240-dimensional PIX, 47-dimensional ZER, and 6-dimensional MOR.
    \item \textbf{MSRC-v1} \footnote{https://www.microsoft.com/en-us/research/project/image-understanding}: This dataset includes 210 image samples from Microsoft Research, categorized into 7 clusters. Each image is depicted through six feature sets: 256-dimensional LBP, 100-dimensional HOG, 512-dimensional GIST, 48-dimensional Color Moment, 1302-dimensional CENTRIST, and 210-dimensional SIFT.
    \item \textbf{Wikipedia} \footnote{http://www.svcl.ucsd.edu/projects/crossmodal}:  This dataset contains 693 documents across 10 clusters, gathered from Wikipedia  articles. Each document is characterized by two feature sets: 128-dimensional WORD and 10-dimensional SIFT.
    \item \textbf{Caltech-5V} \cite{xu2022multi}: A dataset of RGB images, contains 5 views across 1400 instances in 7 clusters: 40-dimensional WM, 254-dimensional CENTRIST, 1984-dimensional LBP, 5412-dimensional GIST, and 928-dimensional HOG.
    \item \textbf{MNIST-USPS} \cite{peng2019comic}: A dataset of 5000 handwritten digits categorized into 10 clusters, where each digit is represented by two feature sets: 784-dimensional MNIST and 784-dimensional USPS.
\end{itemize}

\subsection{Experimental Setup}
\textbf{Comparing methods.}
The methods listed below are utilized for comparative analysis against the proposed approach.

\textit{Single-view traditional clustering methods (the input of these methods is the concatenation of all views)}:
\begin{itemize}
    \item \textbf{KM} \cite{macqueen1967some}: $k$-means clustering, utilizing Euclidean distance for instance comparison.
    \item \textbf{HC} \cite{murtagh2017algorithms}: Applies hierarchical clustering with Euclidean distance and a ward-linkage strategy for an agglomerative process.
\end{itemize}

\textit{Single-view interpretable clustering methods (we perform clustering analysis on each view individually and report the best clustering performance)}:
\begin{itemize}
    \item \textbf{IMM} \cite{dasgupta2020explainable}: This method constructs a threshold tree with $k$ (number of ground-truth clusters) leaves by minimizing the number of mistakes at each node, leads to an approximation ratio close to the  $k$-medians or $k$-means cost.
    \item \textbf{ExKMC} \cite{frost2020exkmc}: This method  starts by constructing a threshold tree having an initial $k$ leaves, allowing the tree to expand to a greater number of leaves according to a user-specified parameter $k'$ (where $k' > k$). In our implementation, $k'$ is set to $2 \times k$, where $k$ denotes the number of ground-truth clusters.
    \item \textbf{Shallow} \cite{laber2023shallow}: This method targets minimizing the 
$k$-means cost function, while incorporating a penalty term in the loss function to encourage the construction of shallow decision trees.
\end{itemize}

\textit{Multi-view clustering methods}:
\begin{itemize}
    \item \textbf{CGL} \cite{li2021consensus}: This approach integrates spectral embedding and low-rank tensor learning within a cohesive optimization framework, fostering mutual enhancement and  learning  a consensus graph within the embedded space. The parameters $\lambda$ and $C$ are both set to 1 and  the nearest neighbor $k$ is set to 15.
    \item \textbf{MFLVC} \cite{xu2022multi}: This method introduces a multi-level feature learning strategy for contrastive multi-view clustering, which separates the reconstruction of low-level view-specific features from the learning of consistent high-level semantics. The parameters are set as follows: $\tau_F=0.5$, $\tau_L=1$.
    \item \textbf{SDMVC} \cite{xu2022self}: This approach leverages global discriminative information to create a consistent target distribution, fostering the learning of distinctive features and uniform multi-view predictions. Moreover, it incorporates an alignment rate mechanism to maintain consistency in multi-view clustering outcomes. Parameters of SDMVC  are set to their default values.
    \item \textbf{MCPL} \cite{cai2024multi}: This method integrates latent and original data insights, initially using pseudo-labels for guidance and then capturing data view complementarities. It includes a latent graph recovery for structural integrity and a refined label fusion technique.
    \item  \textbf{CHOC} \cite{you2024consider}: This approach creates view-specific graphs, differentiating between consistent and specific ones for capturing structural and differential insights. This process culminates in a comprehensive affinity graph for spectral clustering,  optimized by the alternating direction method of multipliers.
\end{itemize}

For our method, the encoders configuration for each dataset is structured as: $Input$-$FC_{128}$-$FC_{64}$ , with fully connected layers denoted by $FC$ and symmetric decoders. Image datasets are converted into one-dimensional vectors for analysis, employing ReLU as the activation function and Adam as the optimizer (learning rate at 0.001). The batch size is set equal to the dataset size.   We specify the number of clusters to be the number of ground-truth clusters and set $e_1$, $e_2$, $maxDep$, $minNum$ and $\lambda$ to 200, 400, 10, 10 and 0.1, respectively.  Standardization is applied to the Mfeat, MSRC-v1, and Caltech-5V datasets to normalize feature scale differences. Additionally, we repeat the clustering ten times for each dataset to derive an average performance evaluation result.

\textbf{Evaluation measures}. To assess clustering performance, we employ the following metrics: 

Purity \cite{rendon2011comparison}: Purity is a clustering evaluation metric that
measures the homogeneity of clusters. It calculates the ratio
of the number of correctly classified data points to the total
number of data points,  defined as follows, with $\Omega=\{\Omega_1,\Omega_2,...\Omega_k\}$  and $\Omega^*=\{\Omega_1^*,\Omega_2^*,...\Omega_m^*\}$ representing the sets of predicted and ground-truth clusters,  respectively:
\begin{equation}
    Purity(\Omega,\Omega^*)=\frac{1}{n} \sum_{i=1}^{k} max_{j}|\Omega_{i} \cap \Omega_{j}^{*}|.
\end{equation}
A high purity score indicates that the
clusters are highly homogeneous and each cluster contains
a single class.

Clustering accuracy (ACC) \cite{cai2005document}: Accuracy is a metric used to evaluate the overall correctness of a clustering model. It measures the proportion of data points that are correctly assigned to their respective clusters compared to the total number of data points, expressed as:
\begin{equation}
    ACC=\sum_{i=1}^{n}\frac{\mathbb{I}(\Omega_i^*, map(\Omega_i))}{N},
\end{equation}
where $map(\Omega_i)$ denotes the permutation mapping function across all potential one-to-one correspondences between clusters and labels. The best mapping can be computed by the Kuhn-Munkres algorithm \cite{lovasz2009matching}. A higher ACC reflects the model's capability to cluster data points into their appropriate clusters accurately.

F1-measure (F1) \cite{assent2008inscy}: F1-measure is a harmonic mean of precision and recall. It is used to evaluate the performance of a clustering algorithm in identifying the relevant data points, which can be defined as:
\begin{equation}
    F1-measure(\Omega,\Omega^*)=\frac{2\times precision \times recall}{precision+recall},
\end{equation}
where 
\begin{equation}
    precision=\frac{TP}{TP+FP}, recall=\frac{TP}{TP+FN}.
\end{equation}
A high F1-measure indicates that the clustering algorithm
has a high precision and recall, and is able to identify relevant data points accurately.
\subsection{Experimental Results}
\renewcommand{\arraystretch}{1.25}
\begin{table*}[htbp]
\centering
\caption{Clustering performance comparison. For each metric across the datasets, the highest score is marked in bold and the second-highest is underlined.}
\label{label1}
\begin{tabular}{l|ccc|ccc|ccc|ccc|ccc}
\toprule
 &\multicolumn{3}{c}{Mfeat} &\multicolumn{3}{c}{MSRC-v1} & \multicolumn{3}{c}{Wikipedia} &\multicolumn{3}{c}{Caltech-5V} &\multicolumn{3}{c}{MNIST-USPS}\\
 \midrule
 Methods &Purity &ACC &F1&Purity& ACC& F1&Purity& ACC& F1&Purity &ACC &F1&Purity& ACC &F1\\
 \midrule
 KM&0.561&0.504&0.498&0.527&0.511&0.405&0.611&\textbf{0.593}&\underline{0.490}&0.498&0.478&0.374&0.776&0.766&0.675\\
 HC&0.568&0.513&0.506&0.457&0.429&0.362&0.609&0.570&0.480&0.451&0.441&0.360&0.836&0.830&0.771\\
 \midrule
 IMM&0.650&0.637&0.522&0.666&0.650&0.535&0.605&0.558&0.470&0.636&0.616&0.467&0.365&0.348&0.246\\
 ExKMC&0.709&0.684&0.587&0.673&0.646&0.542&0.608&0.557&0.473&0.705&0.683&0.547&0.460&0.424&0.305\\
 Shallow&0.683&	0.683&	0.539&	0.730&	0.730&	0.604&	\underline{0.614}	&\underline{0.587}&	\textbf{0.498}&	0.695&	0.695&	0.530&	0.422&	0.383&	0.269\\

 \midrule
 CGL&\underline{0.997}&	\underline{0.997}	&\underline{0.994}	&0.852&	0.847&	0.750&	0.436&	0.382&	0.298&	0.747&	0.712	&0.632&	0.740&	0.698&	0.661
\\
 MFLVC&0.820&	0.820&	0.761&	\textbf{0.914}&	\textbf{0.914}	&\textbf{0.838}	&0.338	&0.338&	0.280&	\underline{0.748}&	\underline{0.748}&	0.631&	\textbf{0.996}	&\textbf{0.996}&	\textbf{0.991}\\

 SDMVC&0.910&	0.910&	0.831&	0.748&	0.729&	0.606&	0.375&	0.375&	0.317&	0.670&	0.670&	0.552&	0.859&	0.838&	0.798\\

MCPL&0.844&	0.831&	0.754&	\underline{0.866}&	\underline{0.866}&	\underline{0.759}&	0.393&	0.354&	0.254&	0.742&	0.729&	0.654&	\underline{0.988}&	\underline{0.988}&	\underline{0.976}\\
CHOC &\textbf{1}&	\textbf{1}&	\textbf{1}	&0.724&	0.714	&0.611&	\textbf{0.619}&	0.557&	0.467	&\textbf{0.832}&	\textbf{0.786}&	\textbf{0.741}&	0.617&	0.613&	0.530\\

 Ours& 0.961&0.961&0.935&0.840&0.838&0.724&0.577&0.545&0.450&\underline{0.748}&0.745&\underline{0.665}&0.661&0.654&0.562\\
 \bottomrule
\end{tabular}
\end{table*}

\begin{figure*}[htbp] 

	\centering  
	\vspace{-0.05cm} 
	\subfigtopskip=2pt 
	\subfigbottomskip=2pt 
	\subfigcapskip=-5pt 
	\subfigure[Average depth]{
		\includegraphics[width=0.48\linewidth]{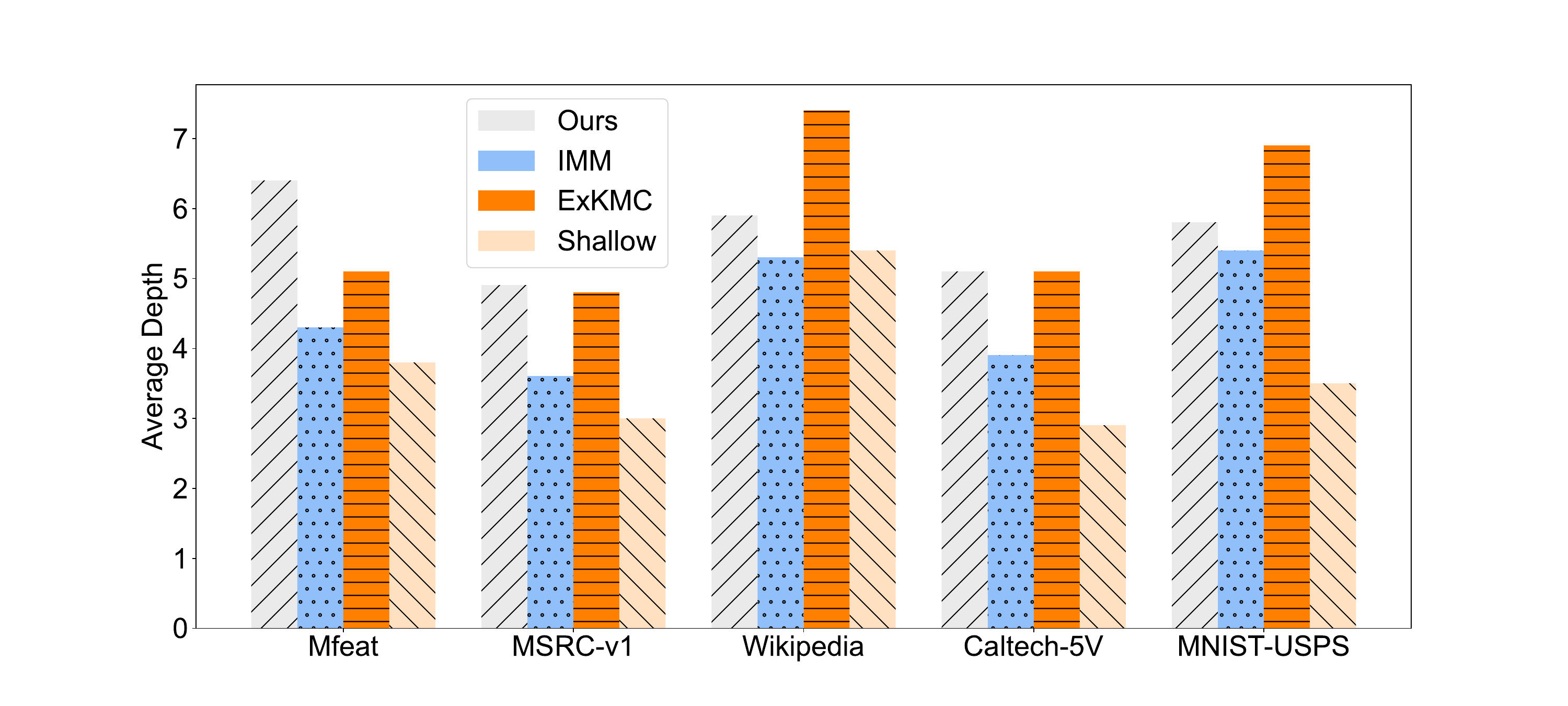}}
	\quad 
	\subfigure[Max depth]{
		\includegraphics[width=0.48\linewidth]{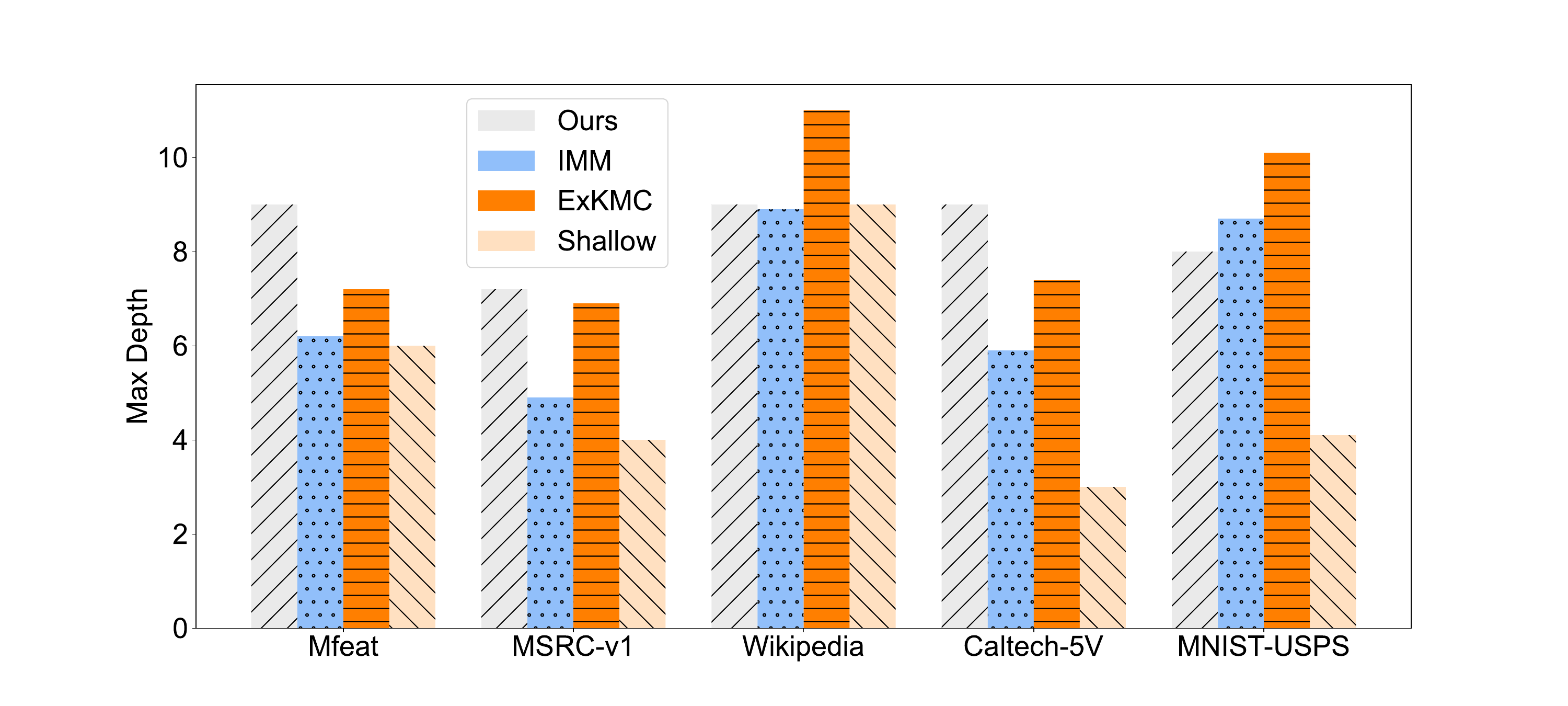}}

	\caption{Comparison of interpretability performance, focusing on the maximum and average depth of  decision trees constructed by different interpretable clustering algorithms.}
 \label{fig2}
\end{figure*}

In Table \ref{label1}, the detailed performance comparison results in
terms of different evaluation metrics are presented. Additionally, the interpretability of decision tree-based clustering methods is quantified by the maximum and average depth of leaf nodes, details of which are presented in Fig. \ref{fig2}. While the number of leaf nodes is also a significant metric for assessing interpretability, it is worth noting that this parameter is predefined for the methods being compared, such as $k$ in IMM and Shallow. Some important observations can be summarized as follows:

\underline{Overall performance}:
 Our method outperforms most compared methods, which demonstrates its superior effectiveness. More precisely, the proposed method can achieve the top three  performance in terms of Purity, ACC and F1-measure on 2 datasets (Mfeat and Caltech-5V). Meanwhile, from Fig. \ref{fig2}, we can find that the interpretability of our method is generally not as good as that of other interpretable clustering methods designed for single-view data.
 
\underline{Comparison with standard clustering methods }:
KM and HC are two popular clustering methods  widely applied in the field of data mining, which can yield comparable results on several datasets. However, our approach generally outperforms them, underscoring its effectiveness in harnessing the complementary information from different views.

\underline{Comparison with interpretable clustering methods}:
In terms of Purity, ACC and F1-measure, our method significantly outperforms IMM, ExKMV and Shallow on almost every dataset (except Wikipedia). However, as indicated in Fig. \ref{fig2}, our method typically results in a lager decision tree compared to trees constructed by other interpretable clustering method, this might because our method opting for a more detailed selection of features to facilitate finer decision tree splits, trading off some level of interpretability to enhance the accuracy of clustering outcomes.

\underline{Comparison with multi-view clustering methods}:
Compared with five SOTA multi-view clustering methods (CGL, MFLVC, SDMVC, CHOC and MCPL),  although our method seldom achieves top-two performance, it still consistently delivers above-average clustering outcomes.  This underscores the strength of our multi-view clustering framework in not only transparently delineating the grouping of data into clusters based on distinct views and features but also in sustaining remarkable accuracy.


%

\subsection{Parameter Sensitivity}
\label{4.4}
In this subsection, we first investigate how the trade-off parameter $\lambda$ influences the clustering performance based on three metrics. As illustrated in Fig. \ref{fig3}, the best clustering results are typically obtained when $\lambda$ is set to 0.1. Overall, the performance across all metrics shows insensitivity to variations in $\lambda$. This insensitivity likely stems from the fact that the final performance is largely dependent on the initially constructed decision tree, highlighting the robustness of the proposed framework.
\begin{figure*}[htbp] 

	\centering  
	\vspace{-0.05cm} 
	\subfigtopskip=2pt 
	\subfigbottomskip=2pt 
	\subfigcapskip=-5pt 
	\subfigure[Purity]{
		\includegraphics[width=0.312\linewidth]{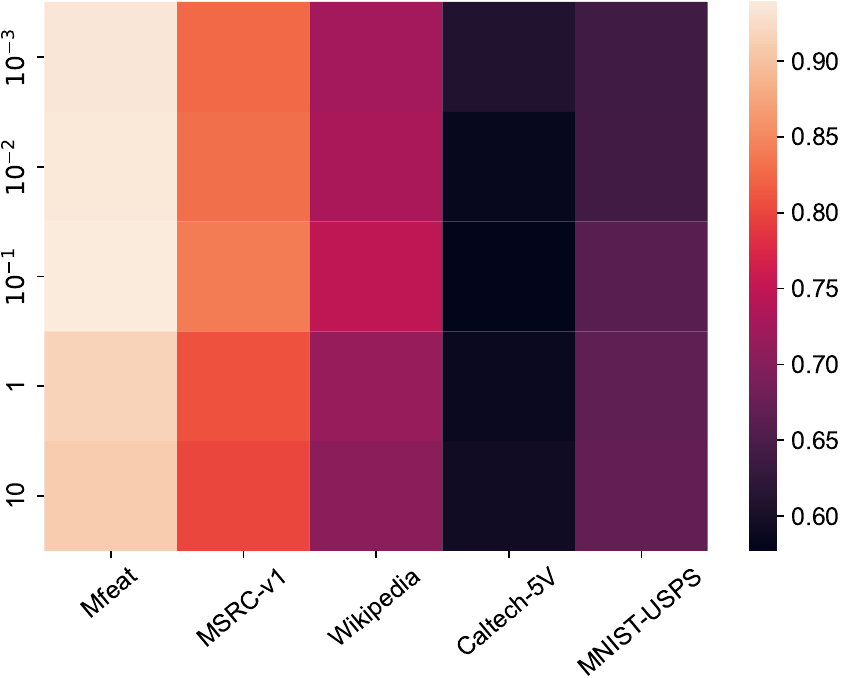}}
	\quad 
	\subfigure[ACC]{
		\includegraphics[width=0.312\linewidth]{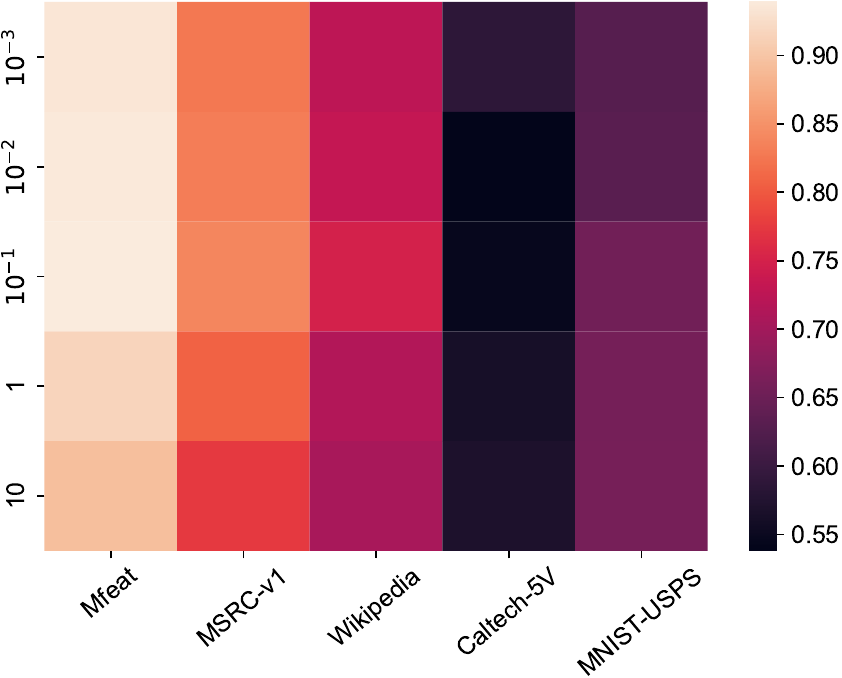}}
        \quad 
	\subfigure[F1-measure]{
		\includegraphics[width=0.312\linewidth]{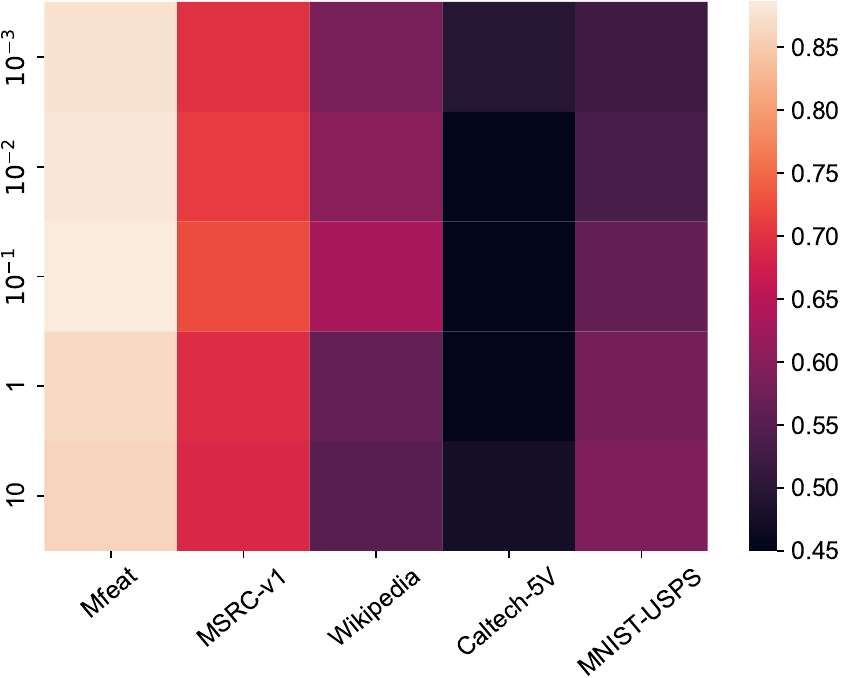}}

	\caption{The effect of parameter $\lambda$ ($y$-axis)  in terms of Purity, ACC and F1-measure.}
 \label{fig3}
\end{figure*}

Secondly, we varied the maximum depth ($maxDep$) of the decision tree from 6 to 10. This parameter influences the size of the decision tree, where a smaller tree generally indicates higher interpretability. From Fig.  \ref{fig4}, it is evident that as the maximum depth of the tree decreases, the performance of the proposed method on three evaluation metrics generally declines. This decrease can be attributed to the decision tree's reduced capability for detailed and precise partition, highlighting the trade-off between interpretability and accuracy in our approach.

\begin{figure*}[htbp] 

	\centering  
	\vspace{-0.05cm} 
	\subfigtopskip=2pt 
	\subfigbottomskip=2pt 
	\subfigcapskip=-5pt 
	\subfigure[Purity]{
		\includegraphics[width=0.312\linewidth]{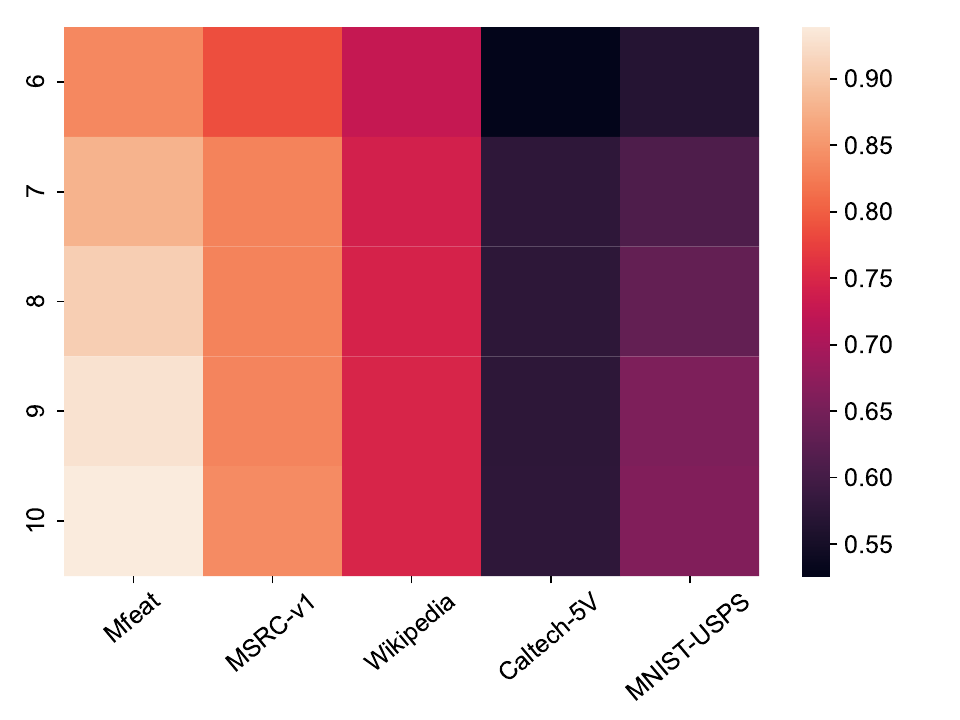}}
	\quad 
	\subfigure[ACC]{
		\includegraphics[width=0.312\linewidth]{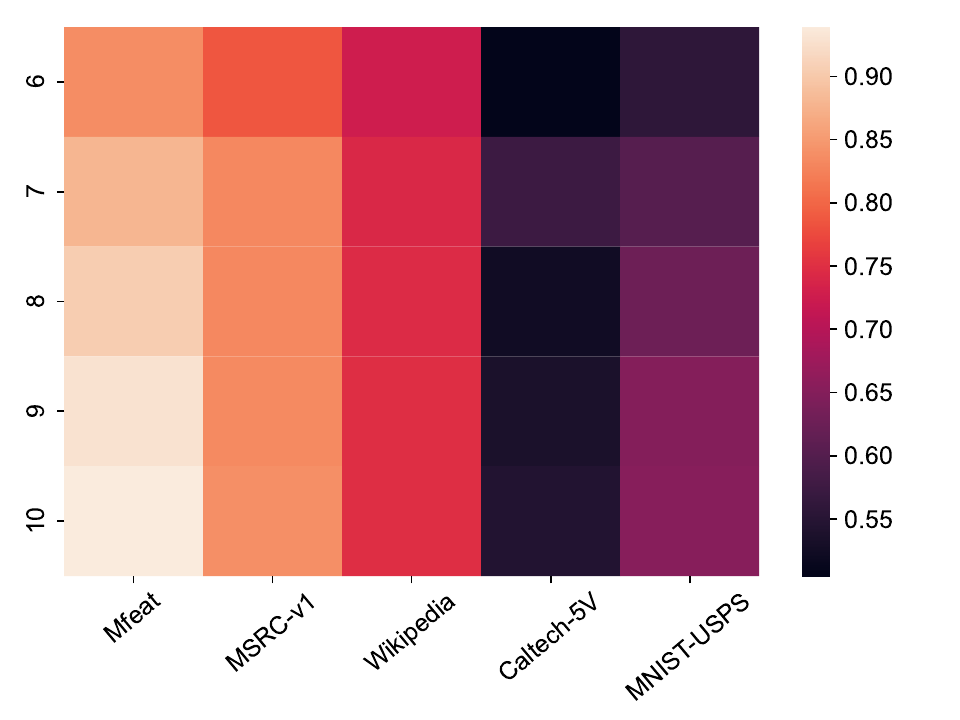}}
        \quad 
	\subfigure[F1-measure]{
		\includegraphics[width=0.312\linewidth]{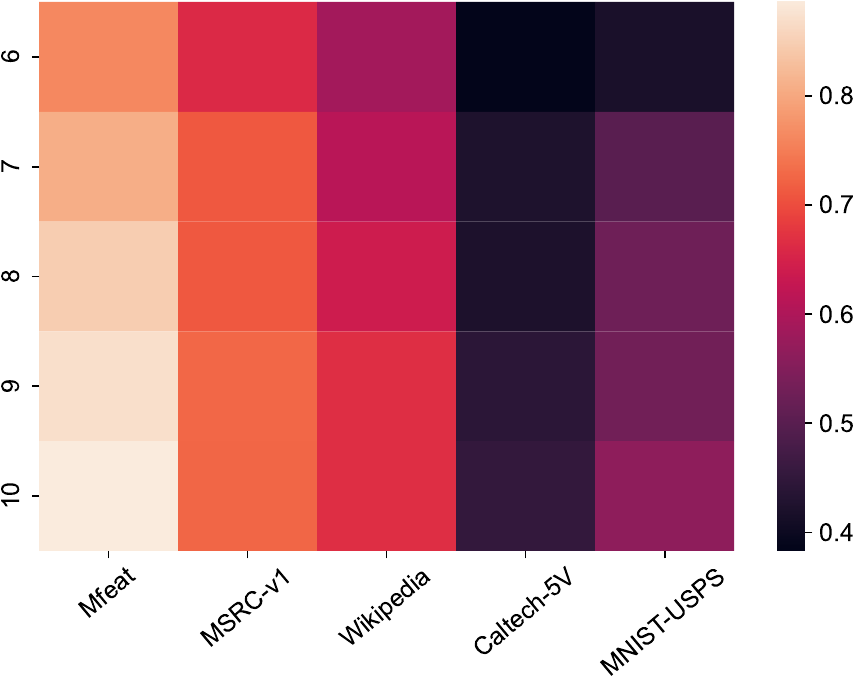}}

	\caption{The effect of parameter $maxDep$ ($y$-axis)  in terms of Purity, ACC and F1-measure.}
 \label{fig4}
\end{figure*}

Finally, the application and design of  interpretable clustering models often lead to a reduction in the accuracy of the final clustering performance compared to initial results. This decrease is typically because the model’s construction and optimization goals focus on fitting the original clustering outcomes as accurately as possible. Therefore, we conduct an experiment by removing the interpretable decision tree, utilizing $k$-means clustering results from concatenated features to establish a consistent data distribution, and then computing the cross-entropy loss with view-specific assignments. The comparative results between the tree-removed model and the full model are presented in Table \ref{table2}, where it is evident that the constructed decision tree adequately fits the clustering outcomes. The average decrease of all performance metrics on all datasets is less than 0.04.

\begin{table}[htbp]
    \caption{Comparison of tree-removed model and full model in terms of average Purity, ACC and F1-measure on all datasets.}
     \label{table2}
    \centering
    \begin{tabular}{c|cc}
    \toprule
     Metric (average) & Tree-removed model   &  Full  model \\
     \midrule
       Purity  & 0.782 &0.753 (-0.029)\\
       ACC  & 0.765 &0.745 (-0.020)\\
       F1-measure  & 0.695 &0.657 (-0.038)\\
        \bottomrule
    \end{tabular}
   
\end{table}
\subsection{The Comparison Via the Visualization of Decision Trees}
In this section, we aim to utilize a dataset of reduced scale and fewer clusters to thoroughly visualize the decision trees constructed by our algorithm compared to other single-view interpretable clustering algorithms. This exercise investigates whether our proposed interpretable multi-view clustering framework can effectively discern features with strong discriminative power to achieve a more precise partition. To accomplish this task, we employ all the data from the first three clusters of the Mfeat dataset, where each cluster corresponds to an unique digit.

\begin{figure*}[htbp] 

	\centering  
	\vspace{-0.05cm} 
	\subfigtopskip=2pt 
	\subfigbottomskip=2pt 
	\subfigcapskip=-5pt 
	
	\subfigure[Ours]{
		\includegraphics[width=0.39\linewidth]{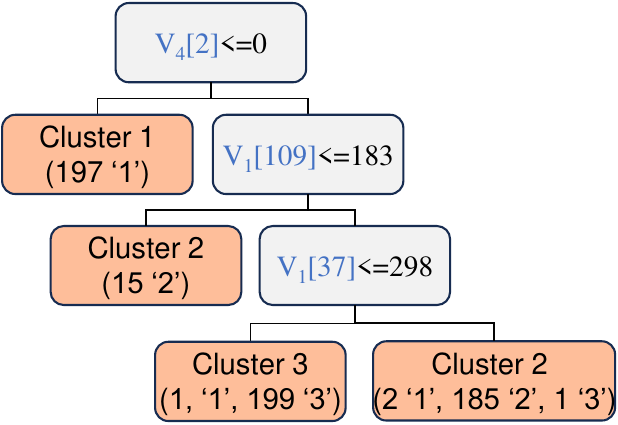}}
         \quad 
        \subfigure[ExKMC]{
		\includegraphics[width=0.52\linewidth]{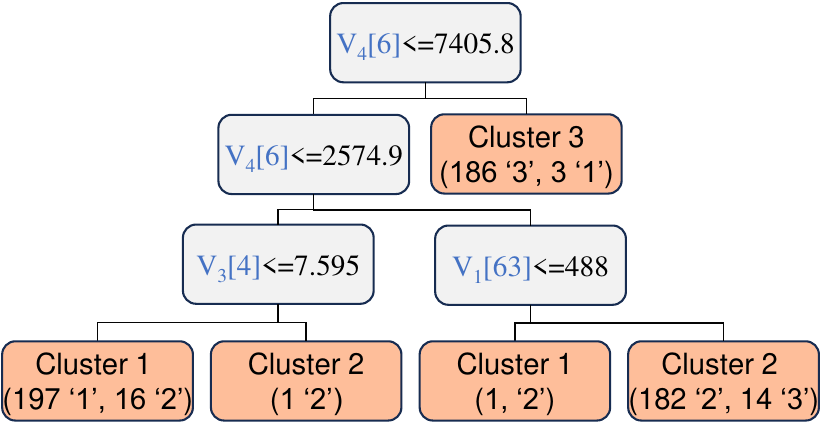}}
	\quad 
	\subfigure[IMM]{
		\includegraphics[width=0.34\linewidth]{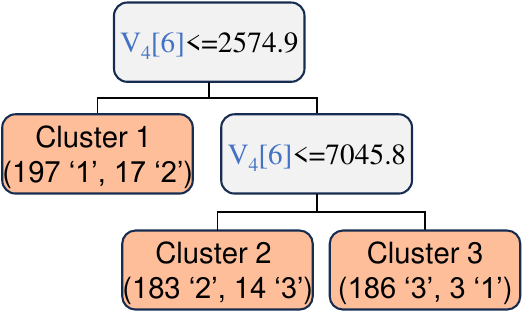}}
        \quad 
	\subfigure[Shallow]{
		\includegraphics[width=0.33\linewidth]{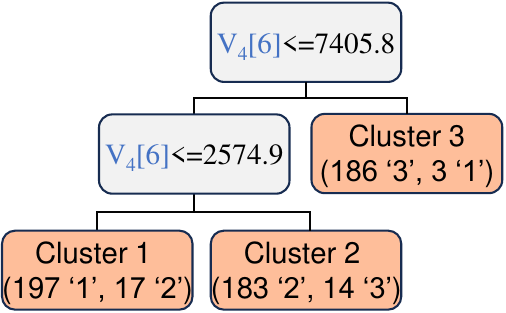}}

	\caption{
Decision trees for interpretable clustering algorithms applied to the Mfeat dataset consisting of three clusters. Here,  $V_1$, $V_2$, $V_3$, $V_4$, $V_5$ and $V_6$ represent features FOU, FAC, KAR, PIX, ZER and MOR of 600 samples, with 200 samples per cluster, respectively. The clusters 1, 2, and 3 correspond to the ground-truth clusters for the digits '1', '2', and '3', respectively.}
 \label{fig5}
\end{figure*}

From Fig. \ref{fig5}, it is observable that although the decision trees constructed by IMM and Shallow have only two internal nodes to complete their construction, resulting in smaller and more interpretable trees, they incorrectly allocate 34 samples at their leaf nodes. In contrast, the decision tree constructed using our method employs an additional internal splitting node, which reduces the number of misclassified samples to 4, significantly enhancing the accuracy compared to the former methods. Additionally, compared to the tree built by ExKMC, our method demonstrates superior accuracy and interpretability. Overall, visualization comparisons show that our method can achieve more precise partitioning at the cost of a slight increase in tree size, confirming the efficacy of the proposed framework.

\section{Conclusion}
\label{5}
In this paper, we present an interpretable multi-view clustering framework that iteratively refine the view-specified feature representation and the interpretable decision tree.  Experimental results on real datasets  demonstrate that our proposed framework not only provides a transparently clustering process for multi-view data  but also delivers performance on par with SOTA multi-view clustering methods.

However, there are still several limitations of our method. First of all, the quality of the constructed decision tree is highly dependent on the pseudo-labels, which leads to limitations in the overall quality of the model. Secondly, as illustrated in Subsection \ref{4.4}, our method struggles to simultaneously balance interpretability and accuracy.

For future work, to address the dependency of the decision tree's quality on pseudo-label accuracy, we may focus on constructing decision trees directly based on the inherent information within the data across different views. Alternatively, the integration of other interpretable models, such as if-then rules, could be contemplated for application to multi-view data.


\appendices

\section*{Acknowledgments}
This work has been supported by the Science and Technology Planning Project of Liaoning Province under Grant No. 2023JH26/10100008, and the National Natural Science Foundation of China under Grant Nos. 62076047, and 61972066. 

\ifCLASSOPTIONcaptionsoff
  \newpage
\fi

\small
\bibliographystyle{IEEEtran}
\bibliography{refj}

\begin{IEEEbiography}[{\includegraphics[width=1in,height=1.25in,clip,keepaspectratio]{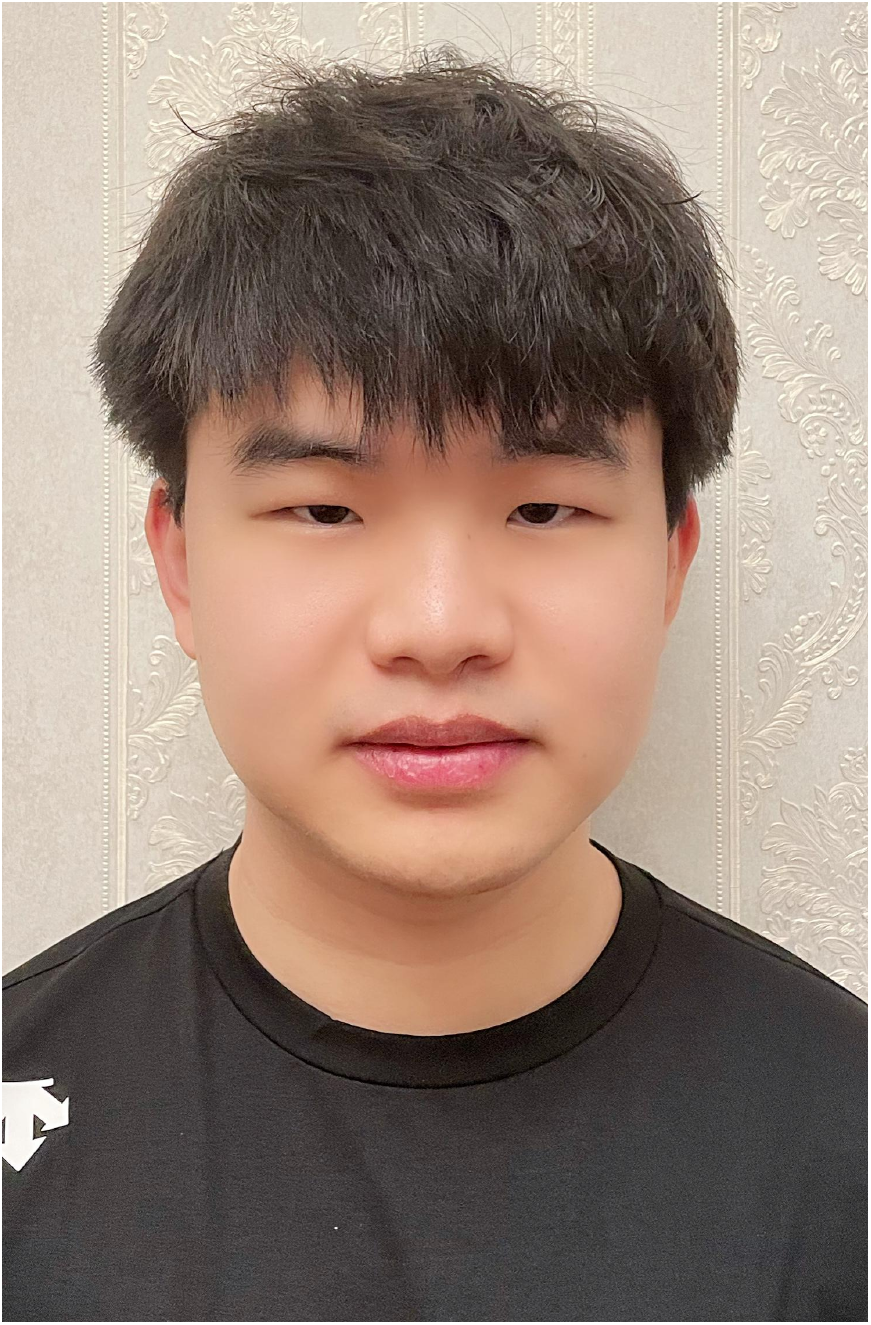}}]{Mudi Jiang}
	received the MS degree in software engineering from Dalian
University of Technology, China, in 2023. He is currently working toward
the PhD degree in the School of Software at the same university. His current
research interests include data mining and its applications.
\end{IEEEbiography}

\begin{IEEEbiography}[{\includegraphics[width=1in,height=1.25in,clip,keepaspectratio]{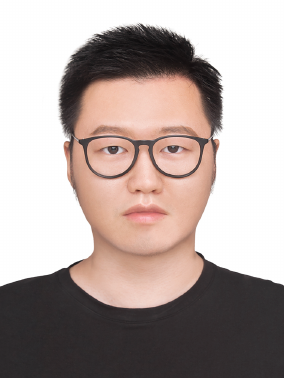}}]{Lianyu Hu}
	received the MS degree in computer science from Ningbo University, China, in 2019.
	He is currently working toward the PhD degree in the School of Software at Dalian University of Technology. His current research interests include machine learning, cluster analysis and data mining.
\end{IEEEbiography}

\begin{IEEEbiography}[{\includegraphics[width=1in,height=1.25in,clip,keepaspectratio]{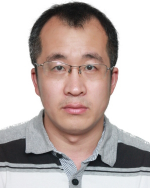}}]{Zengyou He}
	received the BS, MS, and PhD degrees in computer science from Harbin Institute of Technology, China, in 2000, 2002, and 2006, respectively. He was a research associate in the Department of Electronic and Computer Engineering, Hong Kong University of Science and Technology from February 2007 to February 2010. He is currently a professor in the School of software, Dalian University of Technology. His research interest include data mining and bioinformatics.
\end{IEEEbiography}

\begin{IEEEbiography}[{\includegraphics[width=1in,height=1.25in,clip,keepaspectratio]{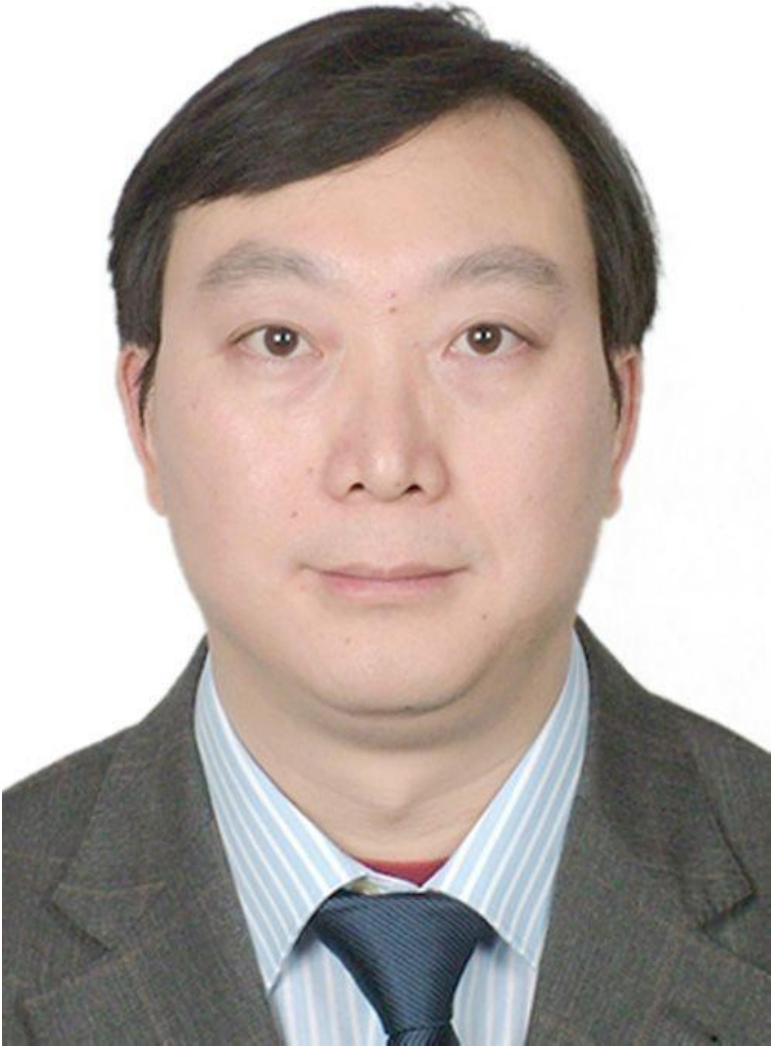}}]{Zhikui Chen}
	(Member, IEEE) received the B.S. degree in mathematics from Chongqing Normal University, Chongqing, China, in 1990, and the M.S. and Ph.D. degrees in mechanics from Chongqing University, Chongqing, in 1993 and 1998, respectively. He is currently a Full Professor with the Dalian University of Technology, Dalian, China.  His research interests are the Internet of Things, big data processing, mobile cloud computing, and ubiquitous networks.

\end{IEEEbiography}


\end{document}